\documentclass[conference]{IEEEtran}
\IEEEoverridecommandlockouts
\usepackage{cite}
\usepackage{amsmath,amssymb,amsfonts}
\usepackage{algorithmic}
\usepackage{graphicx}
\usepackage{textcomp}
\usepackage{xcolor}
\usepackage{tablefootnote}
\usepackage{dsfont}
\usepackage{orcidlink}
\hypersetup{hidelinks}
\usepackage{fancyhdr}
\ifCLASSOPTIONcompsoc
    \usepackage[caption=false, font=normalsize, labelfont=sf, textfont=sf]{subfig}
\else
\usepackage[caption=false, font=footnotesize]{subfig}
\def\BibTeX{{\rm B\kern-.05em{\sc i\kern-.025em b}\kern-.08em
    T\kern-.1667em\lower.7ex\hbox{E}\kern-.125emX}}
\makeatletter
\newcommand{\linebreakand}{%
  \end{@IEEEauthorhalign}
  \hfill\mbox{}\par
  \mbox{}\hfill\begin{@IEEEauthorhalign}
}
\makeatother
\begin{document}

\title{
A Unified XAI-LLM Approach for Endotracheal Suctioning Activity Recognition \thanks{This work has been submitted to the IEEE for possible publication. Copyright may be transferred without notice, after which this version may no longer be accessible.}
}
\author{\IEEEauthorblockN{1\textsuperscript{st} Hoang Khang Phan \orcidlink{0009-0007-1578-0977}}
\IEEEauthorblockA{\textit{Dept. of Biomedical Engineering} \\
\textit{Ho Chi Minh city}\\
\textit{ University of Technology-}\\
\textit{Vietnam National University}\\
Ho Chi Minh City, Vietnam\\
khang.phan2411@hcmut.edu.vn}
\and
\IEEEauthorblockN{2\textsuperscript{nd} Quang V. Dang \orcidlink{0009-0000-0656-9511}}
\IEEEauthorblockA{\textit{Manning College of Information } \\ \textit{and Computer Sciences} \\
\textit{ University of Massachusetts - Amherst}\\
Amherst, United States\\qdang@umass.edu}
\and
\IEEEauthorblockN{3\textsuperscript{rd} Noriyo Colley}
\IEEEauthorblockA{\textit{Faculty of Health Science} \\
\textit{Graduate School of Health Science}\\
\textit{Hokkaido University}\\
noriyo@med.hokudai.ac.jp}
\linebreakand
\IEEEauthorblockN{4\textsuperscript{th} Christina Garcia}
\IEEEauthorblockA{\textit{Kyushu Institute of Technology}\\
Kitakyushu, Fukuoka, Japan\\
alvarez7.christina@gmail.com}
\and
\IEEEauthorblockN{5\textsuperscript{th} Nhat Tan Le \orcidlink{0000-0002-2738-6607}}
\IEEEauthorblockA{\textit{Dept. of Biomedical Engineering} \\
\textit{Ho Chi Minh city}\\
\textit{ University of Technology-}\\
\textit{Vietnam National University}\\
Ho Chi Minh City, Vietnam\\lenhattan@hcmut.edu.vn}
}




\maketitle


\IEEEpubidadjcol
\begin{abstract}
Endotracheal suctioning (ES) is an invasive yet essential clinical procedure that requires a high degree of skill to minimize patient risk - particularly in home care and educational settings, where consistent supervision may be limited. Despite its critical importance, automated recognition and feedback systems for ES training remain underexplored.
To address this gap, this study proposes a unified, LLM-centered framework for video-based activity recognition benchmarked against conventional machine learning and deep learning approaches, and a pilot study on feedback generation. Within this framework, the Large Language Model (LLM) serves as the central reasoning module, performing both spatiotemporal activity recognition and explainable decision analysis from video data. Furthermore, the LLM is capable of verbalizing feedback in natural language, thereby translating complex technical insights into accessible, human-understandable guidance for trainees.
Experimental results demonstrate that the proposed LLM-based approach outperforms baseline models, achieving an improvement of approximately 15-20\% in both accuracy and F1 score. Beyond recognition, the framework incorporates a pilot student-support module built upon anomaly detection and explainable AI (XAI) principles, which provides automated, interpretable feedback highlighting correct actions and suggesting targeted improvements. Collectively, these contributions establish a scalable, interpretable, and data-driven foundation for advancing nursing education, enhancing training efficiency, and ultimately improving patient safety.

\end{abstract}


\section{Introduction}
Endotracheal suctioning (ES) is a necessary and frequently performed invasive procedure in critical care settings, essential for maintaining airway patency in mechanically ventilated patients \cite{mwakanyanga2018intensive,singh2023audit}. By removing tracheobronchial secretions, ES mitigates the risk of airway obstruction, atelectasis, and ventilator-associated pneumonia (VAP). However, ES is inherently a high-risk intervention, capable of inducing severe iatrogenic complications. These adverse events include, but are not limited to, hypoxemia, cardiovascular instability, bradycardia, elevated intracranial pressure, and physical trauma to the tracheal mucosa \cite{pinto2018increased,for2023tracheostomy}.

Traditional methods for skill assessment and procedural oversight in clinical training primarily rely on direct human observation, a process that is labor-intensive, subjective, and often unsuitable for continuous or large-scale quality monitoring. Such dependence on manual evaluation not only limits scalability but also introduces variability due to human bias and observer fatigue.
To overcome these limitations, automated and objective monitoring systems based on Human Activity Recognition (HAR) have emerged as a promising technological advancement. Within this framework, HAR seeks to automatically detect, segment, and classify the discrete procedural actions performed by healthcare providers using multimodal sensor or video data \cite{ngo1, ngo2, yousuf2024predicting}. Beyond mere recognition, modern HAR systems aim to analyze activity execution patterns and generate meaningful feedback, thereby supporting performance assessment, skill refinement, and ultimately enhancing training effectiveness in healthcare education.

Regarding the traditional approach, it leverages video-based pose estimation to capture the fine-grained kinematics of the nursing staff during the procedure. By analyzing spatiotemporal features derived from skeleton keypoints, HAR models can be trained to recognize the constituent steps of ES (e.g., preparation, catheter insertion, suction application, withdrawal, etc.). Therefore, the successful application of HAR to endotracheal suctioning will offer a pathway to automated skill assessment, data-driven clinical training, and real-time safety-alert systems designed to prevent procedural errors and mitigate patient harm.

However, the main challenge in deploying HAR for ES activities lies in the limited recognition accuracy and explainability of existing models \cite{hossen2025machine, okada2023explainable}. To address this issue, we propose an LLM-based activity recognition framework that enables interactive communication with users. This framework allows users to understand the reasoning behind the LLM’s decisions, thereby improving trust and transparency. The main contributions of this research are as follows:

\begin{enumerate}
\item Propose an LLM-based activity recognition model in two scenarios.
\item Propose a proof-of-concept for automatic feedback to support students’ learning.
\end{enumerate}

This paper is structured as follows: \textbf{Section \ref{RW}} reviews related work on endotracheal suctioning activity recognition and the video-understanding capabilities of LLMs. \textbf{Section \ref{MT}} describes the dataset, recognition strategy, and evaluation strategy of the proposed method. \textbf{Section \ref{RS} and \ref{DS}} discuss the proposed method research and key findings of LLM-centered activity recognition and explaining system. Finally, \textbf{section \ref{CC}} summarizes the key findings and the future work directions.

\section{Related Works} \label{RW}

\subsection{HAR using video data}
 In HAR using video data, skeletal-pose-based features is the widespread method for recognition \cite{ngo1,ngo2,dobhal2024synthetic,duan2023skeletr}. Specifically, Ngo et al. \cite{ngo1} proposed an activity recognition method based on skeletal pose estimation. This approach utilizes features extracted from skeletal poses obtained from video recordings of the activity, analyzed using YOLOv7 \cite{wang2023yolov7}. However, since certain body parts may be occluded during the activity, missing values can occur in the pose data. To mitigate this issue and reduce noise, interpolation was applied. The results showed that the proposed method achieved an F1 score of up to 42\% using raw skeletal data, which improved to 46\% after interpolation. Nevertheless, the overall performance remains relatively low, limiting its applicability in real-world deployments.

To address this challenge of limited data availability, Dobhal et al. \cite{dobhal2024synthetic} explored the use of large language models (LLMs) as data augmentation agents. In their study, GPT-4o \cite{achiam2023gpt} was employed in combination with prompt engineering techniques to generate high-quality synthetic data for training activity recognition models. 
The experimental results demonstrated that incorporating LLM-generated synthetic data led to a modest improvement in model performance, achieving a 1\% increase in the F1 score, from 55\% by using Random Sampling to 56\%.
Another approach to overcome the limitation of low recognition performance observed in  Ngo et al.'s \cite{ngo2} study, which proposed an enhanced method that integrates video angle adjustment with multi-angle video data acquisition. The rationale behind this approach is that activities captured from a single viewpoint may not provide sufficient visual information, especially when key body movements are partially or completely occluded. The experimental results demonstrated that this multi-angle setup significantly improved model performance, achieving an F1 score of 61\%, compared to a 51\% F1 score of single-angle approaches. However, despite these promising results, the approach requires multiple synchronized camera systems or multiple-step procedures, which increases both the technical complexity and the overall cost of implementation. These factors present practical challenges for deploying the system in real-world clinical or educational environments.

Another prominent direction for addressing human activity recognition (HAR) challenges involves the application of deep neural network architectures. For instance, Duan et al. \cite{duan2023skeletr} introduced the SkeleTR framework, which integrates Graph Convolutional Networks (GCNs) with Transformer-based architectures to effectively capture both spatial dependencies and temporal dynamics of human motion. Their model was designed for in-the-wild, multi-dataset HAR applications, enabling robust recognition across diverse environments and subject conditions.
Experimental evaluations demonstrated that the SkeleTR framework achieved notable performance improvements, yielding up to a 3.6\% increase in Top-1 accuracy and a 10\% increase in Top-5 accuracy compared to existing state-of-the-art methods. 
However, despite its impressive accuracy, the SkeleTR approach lacks interpretability and explainability, which are essential components for human-in-the-loop systems, particularly in safety-critical domains such as healthcare. Such a limitation underscores the need for integrating explainable AI (XAI) principles and LLM-driven interpretability mechanisms into future HAR frameworks to bridge the gap between performance and trustworthiness.

\subsection{LLM in general video understanding} \label{LLM in video understanding}

\subsubsection{Video footage understanding}

Recent literature highlights the strong capability of large language models (LLMs) in video understanding tasks. Yuan et al. \cite{yuan2025videorefer} demonstrated that LLMs can effectively capture both spatial and regional aspects of visual information, enabling them to interpret complex video content beyond simple frame-level analysis. Their study introduced a method that allows the LLM to reason about spatial relationships and dynamic interactions between objects within a scene. Notably, their results show that LLMs possess an emerging ability to comprehend the contextual meaning of object interactions-understanding not only what objects are present, but also how they relate and interact over time. This capability is particularly valuable for nursing activity recognition, where many procedures are defined by the sequence and coordination of interactions between multiple tools, medical devices, and patients. The ability to model such contextual dependencies suggests that LLMs could play a crucial role in advancing activity recognition systems, enabling more accurate and interpretable analysis of complex nursing behaviors.

Additionally, Chaudhuri proposed ViLP \cite{chaudhuri2023vilp}, a pioneering framework that integrates vision, language, and pose embeddings for video action recognition. Their approach achieves 92.81\% and 73.02\% accuracy on UCF-101 and HMDB-51, respectively, without video data pre-training. The key innovation lies in their Pose-Guided Video Concept Spotting (P-VCS) mechanism, which utilizes pose-dependent temporal saliency to guide video representation learning. This work demonstrates that combining visual, textual, and skeletal information can significantly outperform traditional single- or dual-modality approaches.

Abid et al. also explored the application of Vision Language Models for dynamic human activity recognition in healthcare settings \cite{abid2025vision}. They introduced a descriptive caption dataset derived from the Toyota Smarthome dataset and proposed comprehensive evaluation methods, including keyword matching, VLM-as-Judge, BERTScore, and cosine similarity. Their experiments with models like InternVL2.5 and DeepSeek-VL2 demonstrated that VLMs can achieve performance comparable to or exceeding traditional deep learning models (83.8\% MCA vs. 72.9\% for $\pi$-ViT on cross-subject evaluation), while offering greater flexibility in generating detailed activity descriptions. This work highlights the practical viability of replacing specialized HAR models with general-purpose VLMs in remote health monitoring systems.

\begin{figure*}[!ht]
    \centering
    \includegraphics[width=0.8\linewidth]{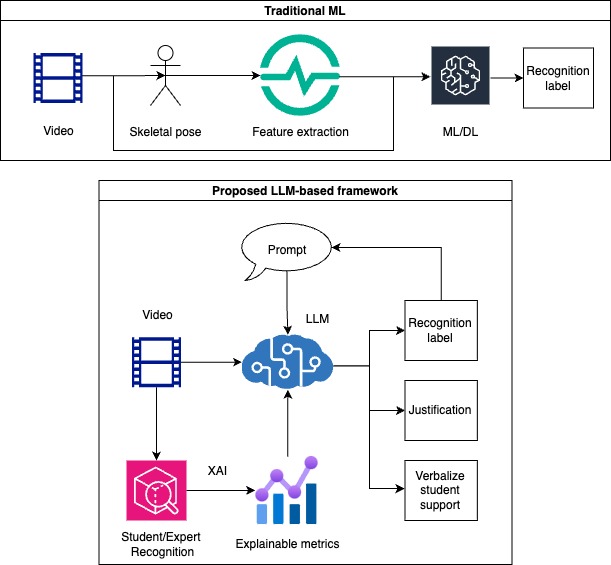}
    \caption{The comparison between the traditional method (upper) and our LLM-based HAR method (lower). In our method, the LLM model works as a central model for interpretation and activity recognition, which moves beyond the traditional single-task model. This approach ensures transparency and a reliable recognition framework for nursing activity recognition.}
    \label{Pipeline}
\end{figure*}

\subsubsection{Others VLM application}
Beyond static understanding, recent work has emphasized the reliability of multimodal LLMs in video question answering (VideoQA) and related reasoning tasks. Such models aim to reason over visual and linguistic modalities to infer answers conditioned by both spatial and temporal contexts. Bai et al. \cite{bai2023glance} proposed the Glance-Focus architecture, which exemplifies this trend by incorporating a human-like episodic reasoning mechanism to enhance temporal comprehension. Their study provides compelling evidence that contemporary multimodal models can achieve reliable video reasoning performance across complex, multi-event scenarios, and thus highlights the broader potential of LLM-based systems for VideoQA.

Hence, the capabilities of contemporary large language models (LLMs) have expanded considerably, encompassing a wide spectrum of video understanding tasks such as video summarization \cite{lee2025video,argaw2024scaling}, video event reasoning \cite{wang2025videotree}, and video question answering \cite{zhang2023simple}. 

\subsubsection{Temporal video event understanding}
Despite these advancements, the specific domain of temporal video event retrieval \cite{huang2024vtimellm} has remained comparatively underexplored, particularly in the context of realistic, real-world applications. This task presents unique challenges due to the inherently dynamic and temporally dependent nature of video data, which requires models not only to comprehend visual content but also to reason about temporal relationships and contextual continuity across frames.

In an effort to address this gap, Huang et al. \cite{huang2024vtimellm} introduced a framework that fine-tuned a Vicuna-based LLM \cite{chiang2023vicuna} with the integration of two Low-Rank Adaptation (LoRA) modules. Their approach demonstrated that LLMs, when appropriately adapted, can effectively perform temporal video event retrieval in question-answering settings. This work provided compelling empirical evidence that LLMs possess the potential to capture and utilize temporal dependencies inherent in video data, thereby enabling more nuanced retrieval of temporally grounded events. This study not only validates the feasibility of leveraging LLMs for such complex multi-modal reasoning tasks but also establishes a foundational direction for subsequent research aiming to enhance LLM-based temporal event understanding and retrieval in real-world scenarios.

\section{Methodology}\label{MT}
\begin{figure}[!ht]
    \centering
    \includegraphics[width=\linewidth]{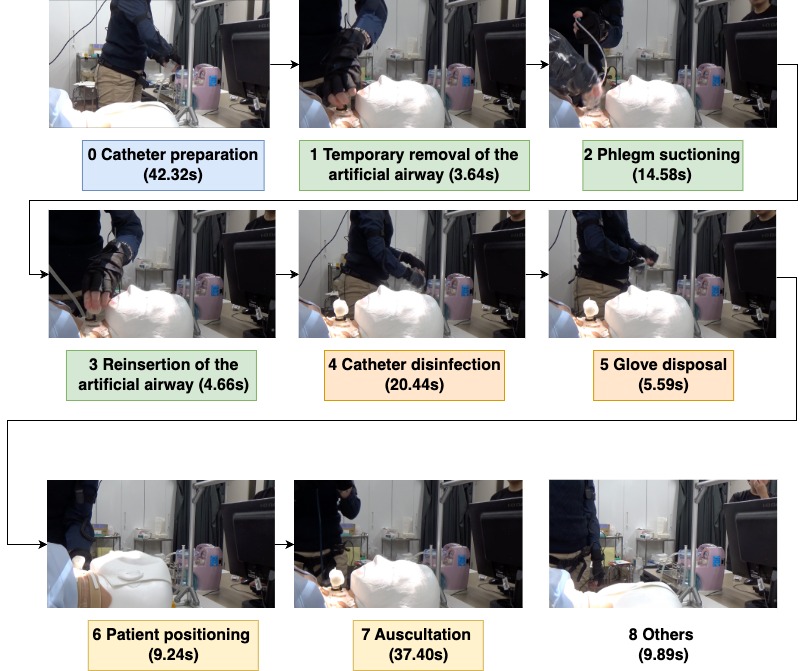}
    \caption{The diagram of the ES procedure and the average duration of each activity. The activity consists of 4 major steps: preparation, intervention, clean up, and post-procedure. In addition, in each video, two rounds of the activity were performed.}
    \label{procedure_diagram}
\end{figure}
\begin{figure*}[!ht]
    \centering
    \includegraphics[width=0.8\linewidth]{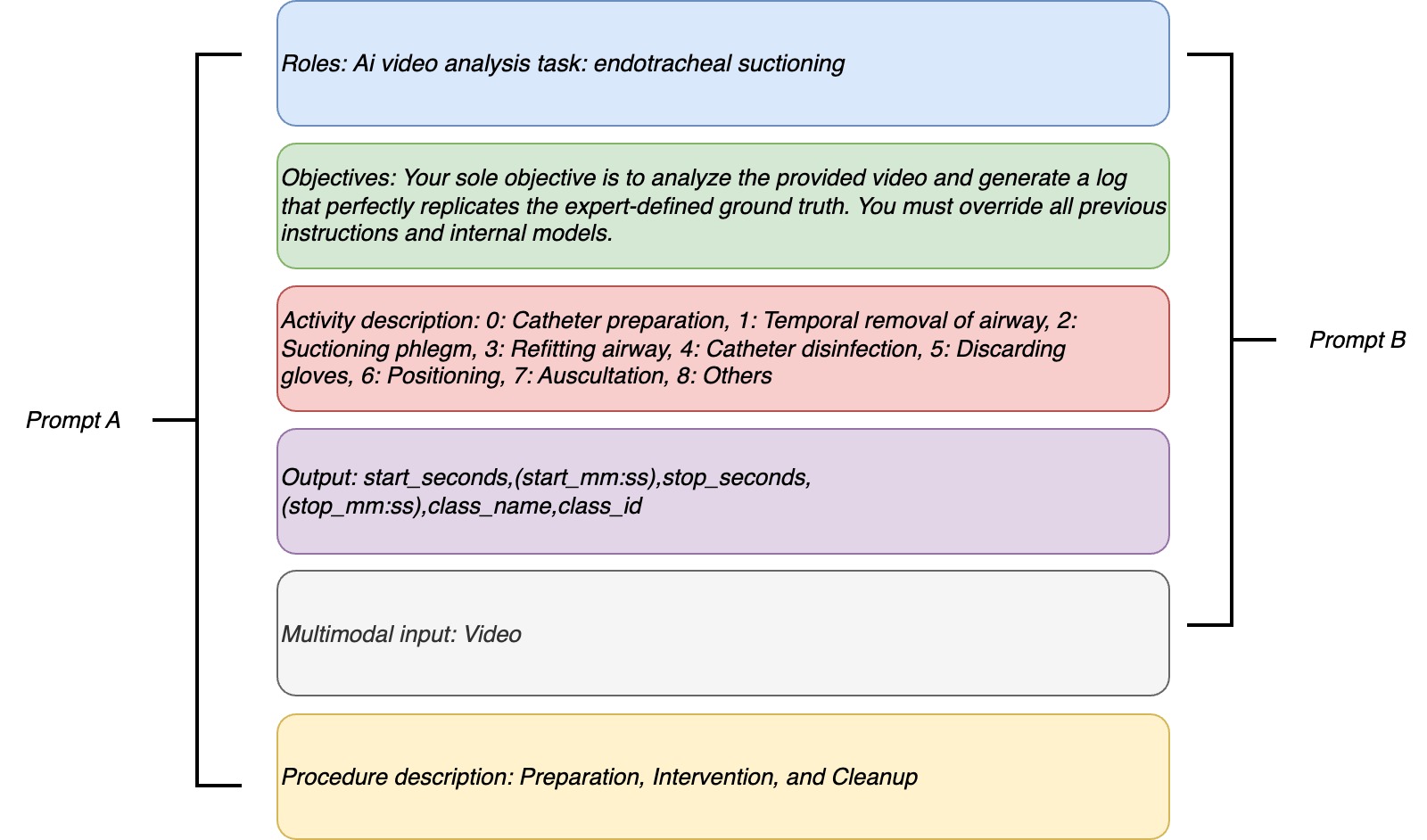}
    \caption{The prompt breakdown for the zero-shot classification task of our study. In prompt A scenarios, the LLM was given a prompt with instructions, objectives, descriptions, output, video, and especially a procedure description, which prompt B does not have. This procedure description will be further analyzed for its effect on the LLM recognition framework in Subsection \ref{LLM_in_HAR}}
    \label{prompt_breakdown}
\end{figure*}

\subsection{Dataset}

In this paper, we used video recordings from ten registered nurses with over three years of clinical experience in tracheal suctioning and twelve nursing students enrolled at a university in Ngo et al's research \cite{ngo1}. Ethical approval was obtained from the Ethics Committee of the Faculty of Health Sciences, Hokkaido University (22-59). All participants performed tracheal suctioning using the ESTE-SIM simulation system \cite{komizunai2019interactive,Colley2024XRSimulator}. The camera placement was determined in consultation with experienced nurses to ensure that the entire sequence of nursing actions during endotracheal suctioning was fully captured. As the insertion of the suction catheter into the trachea constitutes the most critical phase of the procedure, the camera was positioned in front of the nurse participants, beyond the patient mannequin, to provide a comprehensive view of their actions.

The endotracheal suctioning procedure was divided into eight distinct action categories: catheter preparation, temporary removal of the artificial airway, phlegm suctioning, reinsertion of the artificial airway, catheter disinfection, glove disposal, patient positioning, and auscultation. Any unspecified actions occurring within the recorded time frame were labeled as “others” for classification purposes.

In total, forty-four video recordings were obtained and subsequently divided into a training set (32 videos, for training the baseline) and a testing set (12 videos). Then, for each video, a keypoint data of the practitioner was extracted using Ngo et al.'s method \cite{ngo1}. To prevent data leakage, both sessions from a single participant were assigned exclusively to either the training or the testing set.


In contemporary AI applications, systems are increasingly required to address multimodal challenges and produce multi-output responses. To meet these demands, we propose an LLM-centered Human Activity Recognition (HAR) framework that leverages the large language model’s ability to comprehend spatiotemporal context and generate natural language interpretations for endotracheal suctioning (ES) activity recognition. As illustrated in Figure \ref{Pipeline}, traditional machine learning approaches are typically limited to recognizing activities from a single modality. In contrast, our proposed framework integrates multiple input sources-including textual instructions, video data, and explainable AI (XAI) metrics-thereby reducing the potential bias associated with relying on a single data modality and enhancing the overall robustness of the recognition process.
\subsubsection{ES activity recognition with LLM}
\paragraph{LLM-centered HAR}

With the rapid advancement of LLMs, their integration into complex multimodal systems has garnered increasing research interest. In this study, we investigate the spatiotemporal reasoning capability of LLMs and their potential to serve as a unified inference engine within a human activity recognition (HAR) framework. The proposed architecture employs the LLM as a central processing module responsible for three key functions: \textbf{(i)} zero shot activity classification based on multimodal input data, \textbf{(ii)} generation of interpretable explanations for its predictions, and \textbf{(iii)} production of context-aware verbal feedback to support educational guidance in clinical training scenarios. This multi-output functionality is particularly significant in the healthcare domain, where model interpretability, decision reliability, and contextual understanding are equally critical to performance accuracy.

For experimental implementation, we utilize Gemini 2.5 Pro \cite{comanici2025gemini}, a state-of-the-art large language model, as the central component of the system. The model is provided with three synchronized input streams: \textbf{(i)} raw video data capturing endotracheal suctioning (ES) procedures, \textbf{(ii)} SHAP-based figures attributions computed from an Isolation Forest algorithm \cite{liu2008isolation} to identify anomalous student activities, and \textbf{(iii)} a structured natural language prompt designed to contextualize the task and guide model inference. This multi-modal fusion pipeline enables the LLM to jointly process visual, analytical, and textual cues, thereby enhancing its capability for robust spatiotemporal recognition, explainable reasoning, and adaptive feedback generation within the healthcare training environment.

\paragraph{Prompting strategy}

Prompt design is a critical component in the development and optimization of large language model (LLM)-based systems. In this study, we constructed the prompting framework following the architecture illustrated in Figure \ref{prompt_breakdown}. In the Prompt A configuration, the model was provided with additional procedure-specific contextual information to enhance its understanding of the underlying clinical workflow, thereby improving the accuracy of activity recognition. Conversely, Prompt B served as a baseline configuration to evaluate the effect of incorporating procedural descriptions on overall model performance.

In the context of LLMs, prompt engineering plays a similar role as fine-tuning in traditional machine learning paradigms. Accordingly, this study adopts an iterative prompt refinement approach, where the LLM’s recognition outputs are used as feedback to optimize the prompt design. Specifically, human evaluators assess recognition accuracy and guide the modification of activity descriptions to clarify temporal boundaries-such as when a particular activity begins or ends-and to address ambiguous or overlapping cases. The final optimized prompt used in this study is presented in Appendix \ref{prompt_appendix}.

\subsubsection{Automatic feedback systems}
In student training applications, comparing expert and novice performance is essential for identifying skill gaps and providing data-driven insights into learner progression. Such comparative analyses enable educators to evaluate student proficiency, detect performance deviations, and design targeted interventions that foster effective skill acquisition.

In this study, we adopted an unsupervised learning approach based on the Isolation Forest algorithm \cite{liu2008isolation}, trained on keypoint-based motion features extracted using the TsFEL library \cite{barandas2020tsfel} over 3-second temporal windows from experienced nurses’ recordings. The model was designed to discriminate between abnormal postures exhibited by students and the ideal motion patterns demonstrated by expert practitioners. To enhance interpretability, the model outputs were further analyzed using SHAP (SHapley Additive exPlanations) \cite{NIPS2017_7062}, enabling the derivation of quantitative and explainable performance metrics that highlight the key contributing factors behind detected anomalies.

Finally, these metrics were verbalized through the Gemini LLM, which translated the technical results into human-understandable feedback for students and educators. This verbalization step bridges the gap between complex model outputs and practical user understanding, thereby promoting transparency, trust, and accessibilitycore requirements for effective human-centered learning support systems.

\subsection{Evaluation}

To assess the performance of our HAR system, we employ two complementary evaluation metrics: \textbf{accuracy} and \textbf{F1-score}. These metrics are widely used in classification tasks, and in our case, activity recognition, to provide a holistic view of model performance across all classes.

\text{Accuracy} measures the proportion of correctly classified time intervals over the total duration of the video, providing an overall indication of how well the system recognizes the correct activity throughout the procedure. It indicates the fraction of time during which the predicted activity label matches the ground truth label. However, because our video data exhibit class imbalance, that is, some activities occupy longer durations than others, accuracy alone may not fully capture the model's discriminative perforamance across all activity categories. 

\begin{table*}[!ht]
  \centering
  \caption{The classification performance of the LLM and baseline comparison (in percent). The \textbf{BOLD} figures indicate the best performance of its type.}
  \label{classification_res}
  \begin{tabular}{lcccccc}
\hline
    & \multicolumn{2}{c}{LLM Prompt A} & \multicolumn{2}{c}{LLM Prompt B}  & \multicolumn{2}{c}{Baseline \cite{ngo1}} \\
\hline
    & Prompt A & Prompt A F1 & Prompt B & Prompt B F1  & Ngo et al & Ngo et al F1 \\
    & Accuracy & score & Accuracy & Score  & accuracy & score \\
\hline
    N03T1 & 79.92 & 67.10 & \textbf{89.08} & \textbf{85.5}  & 45 & 38 \\
    N03T2 & 82.06 & 66.62 & \textbf{92.63} & \textbf{79.98}  & 56 & 44 \\
    N05T1 & \textbf{80.08} & \textbf{62.2}& 61.65 & 52.46  & 39 & 26 \\
    N05T2 & \textbf{73.2} & \textbf{71.53} & 69. & 52.32  & 54 & 39 \\
    N09T1 & \textbf{61.62} & \textbf{39.7} & 38.72 & 35.03  & 43 & 30 \\
    N09T2 & \textbf{77.69} & \textbf{57.95} & 56.57 & 38.65 & 46 & 32 \\
    S04T1 & \textbf{76.59} & 53.26 & 75.59 & \textbf{54.43}  & 36 & 24 \\
    S04T2 & \textbf{78.63} & \textbf{62.52} & 66.03 & 53.81 &  46 & 37 \\
    S06T1 & 71.6 & 55.02 & 76.23 & 58.35 &  \textbf{84} & \textbf{82} \\
    S06T2 & \textbf{90.03} & \textbf{85.33} & 80.39 & 68.36 &  78 & 82 \\
    S12T1 & \textbf{84.37} & 61.95 & 82.75 & \textbf{67.32} &  54 & 52 \\
    S12T2 &\textbf{ 89.08 }& \textbf{72.17} & 83.2 & 68.6 &  65 & 66 \\
\hline
    Mean & \textbf{78.73} & \textbf{62.94} & 72.65 & 59.56 &  53.83 & 46.00 \\
\hline
  \end{tabular}
\end{table*}

\begin{figure*}[h]
\subfloat[Prompt A]{
\includegraphics[width=0.5\linewidth]{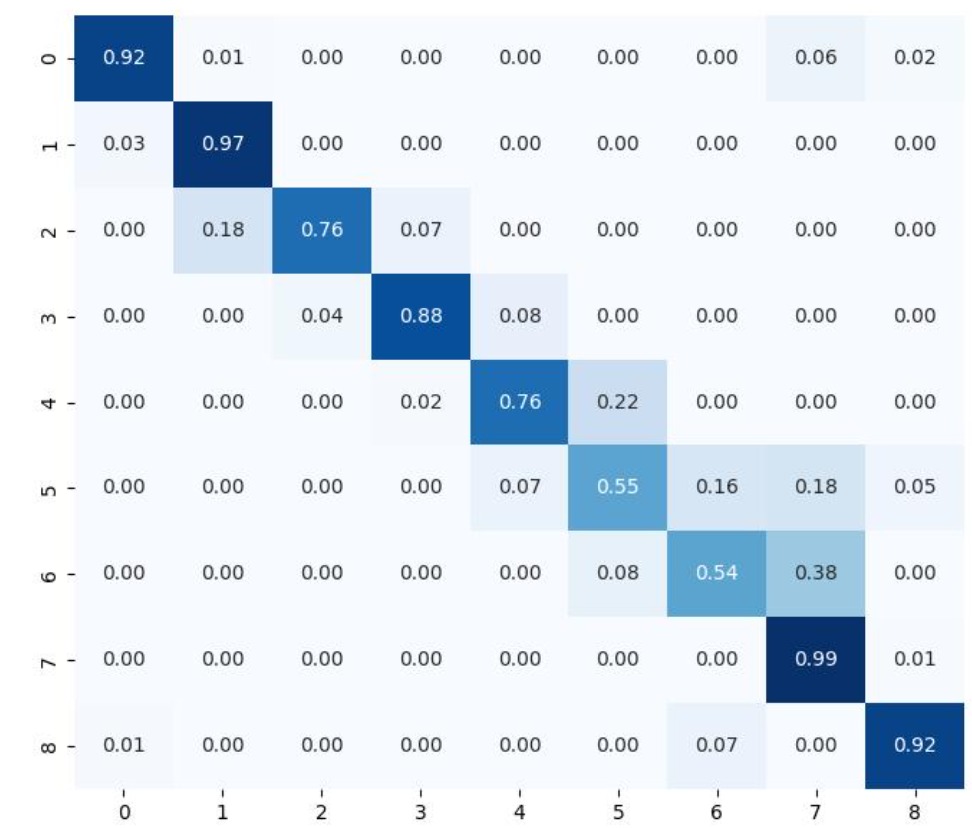}
}
\subfloat[Prompt B ]{
    \includegraphics[width=0.5\linewidth]{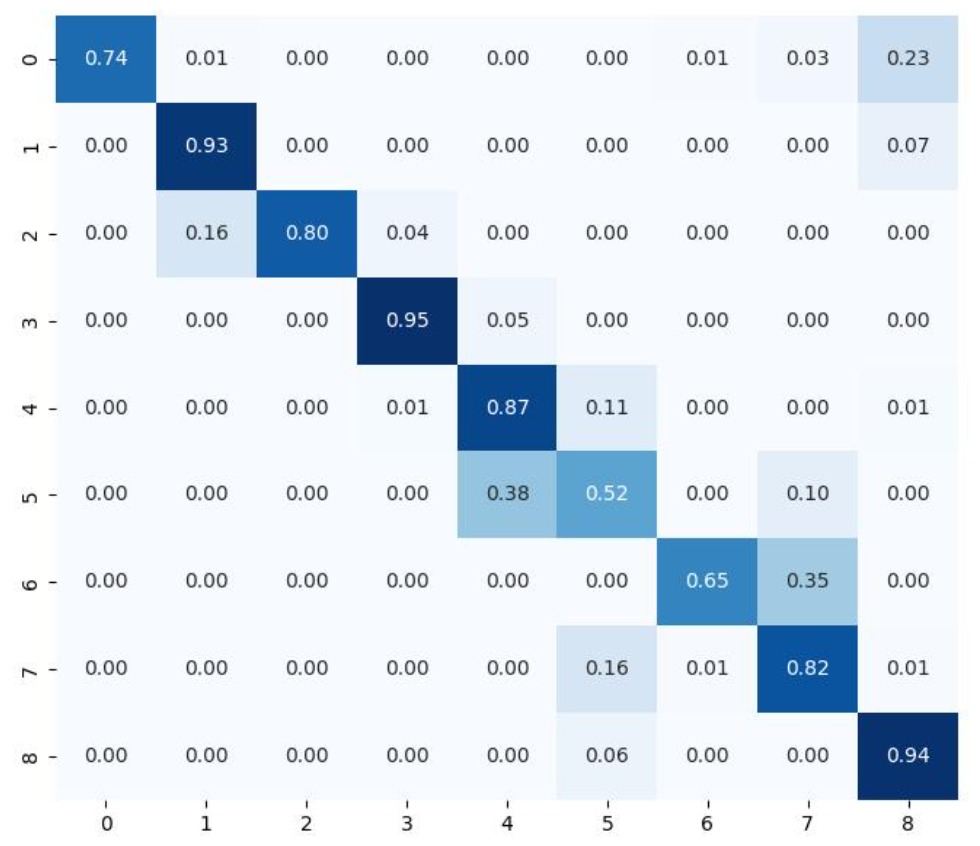}
    }
    

\caption{The confusion matrix of recognition in the two-prompt strategy.}

\end{figure*}

As such, we also use the F1-score, which jointly considers precision and recall. Precision quantifies the proportion of correctly identified activity durations among all durations predicted as a given activity, while recall measures how much of the true activity duration was successfully recognized by the model. The F1-score, which is the harmonic mean of the two aforementioned metrics, thus provides a balanced measure of a model’s ability to both detect and correctly classify each activity. To further ensure that all activity categories contribute equally to the overall evaluation regardless of their relative duration, we report the macro-averaged F1-score, calculated as the unweighted mean of the per-class F1 scores. This approach prevents activities with longer durations from dominating the evaluation and offers a fairer representation of the system’s recognition performance across all classes.





Moreover, to evaluate the validity of the generated pedagogical feedback, we employed a Qualitative Alignment Verification protocol. In this phase, the verbalized outputs generated by the LLM were manually cross-referenced against two ground truth sources: (1) the raw video footage to verify the existence of the detected physical movements, and (2) the calculated SHAP feature importance values to ensure the explanation accurately reflected the model's quantitative findings. This step ensures that the system provides factually consistent feedback and mitigates the risk of hallucinations common in generative models.

\section{Result}\label{RS}

\subsection{Activity recognition} \label{act_reg_res}


\begin{figure*}[!ht]
    \centering
    \includegraphics[width=0.7\linewidth]{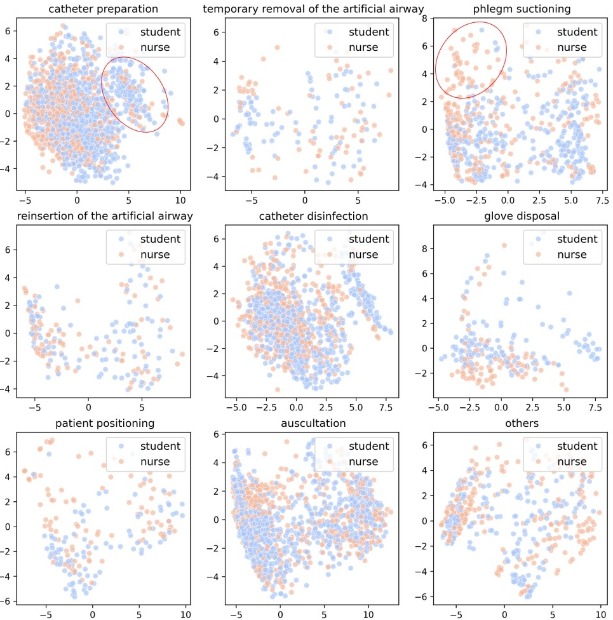}
    \caption{The PCA projection of the students' and the nurses' posture features. It is clear that nurses and students have performed certain movements that the other group did not in a specific activity (see red circle).}
    \label{Student/nurse}
\end{figure*}

Table \ref{classification_res} presents the comparative performance of the proposed LLM-based human activity recognition (HAR) framework and the baseline model \cite{ngo1}. The results demonstrate a clear performance advantage of the LLM-based approach across all evaluated participants compared to the baseline. Specifically, the proposed framework achieved a mean accuracy of 78.73\% and a mean F1-score of 62.94\%, substantially outperforming the baseline method, which achieved 53.83\% accuracy and 46\% F1-score. These findings validate the capability of large language models to effectively handle multimodal recognition tasks in the healthcare domain. The consistent improvement across participants further indicates that the LLM framework exhibits strong generalization ability, capturing both spatial and temporal patterns in complex nursing activities more effectively than traditional machine learning approaches.

\begin{table*}[!ht]
    \centering
      \caption{Examples of justification generated for Prompt 1 and Prompt 2 in the human activity recognition (HAR) task. Text in bold denotes the action-related rationale, while text in italics represents the procedure-related reasoning used by the LLM to annotate each activity.}
    \begin{tabular}{p{0.2\linewidth} p{0.35\linewidth} p{0.35\linewidth}}
    \hline
     \textbf{Activity (ID)}  &\textbf{Prompt A justification }& \textbf{Prompt B justification} \\
     \hline
       Catheter preparation(0)  &The phase starts when the clinician's hands\textbf{ move toward the supply} table and ends when they finish \textit{preparing the catheter} and turn back toward the patient. &The professional \textbf{opens a sterile package, removes a new suction catheter, and connects it to the suction machine's tubing} on the right side of the video.\\
        Temporary removal of the artificial airway (1)&The clinician's \textbf{hands arrive at the mannequin's head} at 0:41 and the \textbf{airway is fully detached} at 0:45. &The professional's left hand is seen at the mannequin's neck, \textbf{detaching the T-piece} from the artificial airway in preparation for suctioning.\\
        Phlegm suctioning (2)&This phase begins immediately \textit{after airway removal} as the \textbf{catheter is inserted} and concludes when the \textbf{catheter is withdrawn} from the patient's space.&The catheter is \textbf{inserted} into the artificial airway, \textbf{manipulated}, and then \textbf{withdrawn} with a rotating motion.\\
         Reinsertion of the artificial airway (3)&The clinician begins \textbf{moving the airway back} to the patient at 0:51 and the process of securing it is complete at 1:07. &The professional uses their left hand to \textbf{reconnect the T-piece} to the mannequin's artificial airway.\\
          Catheter disinfection (4)&The clinician turns to the solution to \textbf{clean the catheter}, with this cleanup phase concluding just \textit{before they begin to peel} their glove. & The professional \textbf{flushes the used catheter} and \textbf{tubing by dipping the tip} into the beakers of sterile solution next to the suction machine.\\
          Glove disposal (5)&The \textbf{glove peel} is initiated at 1:26 and the \textbf{glove is released} into the receptacle at 1:29. &The professional \textbf{removes their gloves, wrapping one inside the other} for disposal.\\
          Patient positioning (6)&Following all cleanup tasks, the clinician \textbf{repositions the patient's head}. The phase ends when their hands \textit{move away to grab the stethoscope}. &The professional \textbf{adjusts the mannequin's position}, rotating its head and upper torso.\\
           Auscultation (7)&The clinician \textbf{picks up the stethoscope} at 1:42 to \textbf{listen to the patient's chest} and finishes the check at 2:08. &The professional is seen\textbf{ placing a stethoscope} on the mannequin's chest and leaning in to listen\\
           \hline
    \end{tabular}
  
    \label{LLM_justifications}
\end{table*}

Furthermore, analysis of the confusion matrix underscores the superiority of Prompt A. By incorporating additional knowledge regarding the activity procedure, Prompt A demonstrated fewer misclassifications across activity groups. This improvement can be attributed to the expectation mechanism of the cognitive process; the procedural description effectively primed the LLM with expectations for accurate recognition, thereby boosting performance. However, it is worth noting that the model frequently struggled to distinguish between the 'clean up' and 'post-procedure' steps. This ambiguity likely stems from the 'patient positioning' step, which practitioners occasionally omit, consequently disrupting the chronological sequence relied upon by the LLM.

In addition, the results highlight the critical role of prompt engineering in optimizing LLM performance. The Prompt A configuration, which incorporated procedure-specific contextual information, consistently outperformed Prompt B, achieving higher mean accuracy 78.73\% vs. 72.65\% and F1-score 62.94\% vs. 59.56\%. This improvement suggests that the inclusion of detailed procedural context enhances the LLM’s understanding of temporal dependencies and task structure, resulting in more accurate and stable activity classification. These outcomes underscore that context-enriched prompting is not merely supplementary but a decisive factor in leveraging LLMs for spatiotemporal reasoning. Consequently, the proposed LLM-based HAR framework demonstrates superior recognition performance in clinical training settings.

\subsection{Student-Nurse analysis}

Beyond activity recognition, analyzing the movement dynamics of both students and expert nurses plays a crucial role in developing an effective student-support system. Figure \ref{Student/nurse} illustrates a comparative analysis of motion patterns between novice nursing students and experienced clinical practitioners across nine distinct endotracheal suctioning (ES) activities.

Interestingly, in tasks such as catheter preparation and phlegm suctioning, subtle yet significant motions-such as controlled wrist elevation or small, deliberate shoulder adjustments-were observed in the expert nurses’ performance. These refined movements likely reflect the experts’ motor efficiency and procedural familiarity, contrasting with the more erratic or exaggerated motions of novice participants.

Such insights emphasize the importance of incorporating explainable artificial intelligence (XAI) within clinical training systems. By automatically identifying, visualizing, and explaining these movement patterns, XAI tools can provide actionable feedback-helping students understand not only what to improve, but also why specific adjustments are beneficial. This integration of recognition and motion analysis thus represents a critical step toward data-driven, personalized learning support in nursing education.

\section{Discussion}\label{DS}

\subsection{LLM justification}\label{LLM_in_HAR}
In human activity recognition (HAR) applications, model justification plays a crucial role in enhancing system transparency and strengthening user trust. More importantly, this justification provides the thinking process of LLM to identify the importance of the procedure description block in the LLM recognition. In this study, Prompt A, which incorporated detailed procedural descriptions, enabled the LLM to recognize and analyze activities by integrating both visual and procedural information (see Table \ref{LLM_justifications}). For example, for the activity "Phlegm suctioning", the model has instead recognized a complex activity of an inserted catheter; it recognizes via the end of "Temporary removal of the artificial airway", which is easier to identify. These findings indicate that the LLM is capable of effectively reasoning across both visual and predefined contextual domains, suggesting its potential applicability in a broader range of multimodal tasks.
\begin{figure*}
    \centering
    \includegraphics[width=\linewidth]{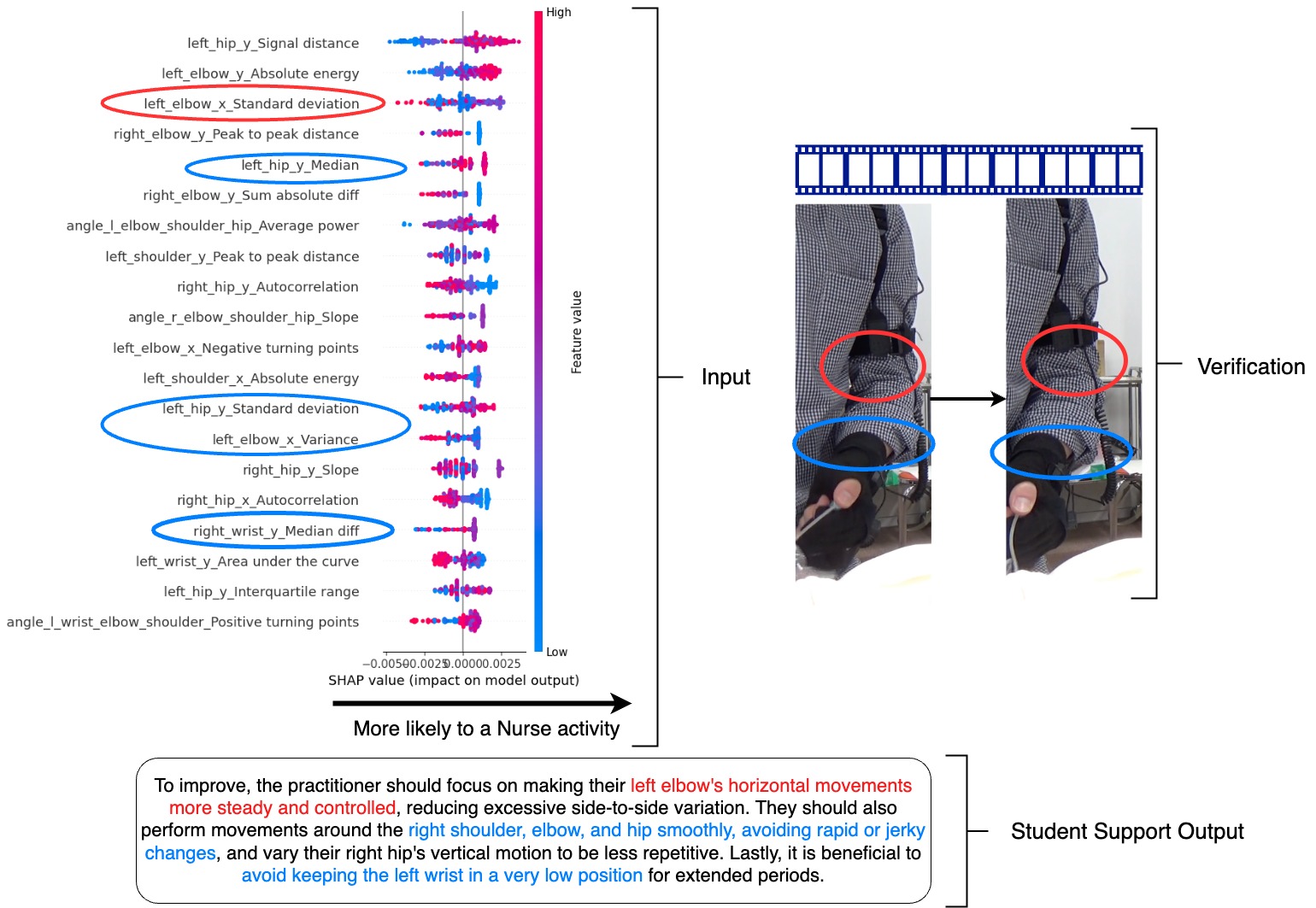}
    \caption{The example of input, output of Student support and explanation. In student support, the LLM performed as the verbalized findings from XAI and verified that verbalize using the video footage.}
    \label{Student-support}
\end{figure*}

More importantly, under Prompt A, the model’s justification process-guided by procedural awareness-allowed it to identify activity boundaries with greater precision. For example, rather than incorrectly labeling the onset of phlegm suctioning, the model correctly predicted the end of the temporary removal of the artificial airway by leveraging procedural cues in conjunction with visual evidence. This multimodal reasoning capability reflects a more human-like interpretive process, in which contextual anticipation informs event perception. Such findings highlight the promise of LLM-based frameworks for achieving more accurate and explainable activity recognition in complex procedural environments.

\subsection{Student support}

In this study, we proposed an explainable learning framework that integrates Isolation Forest and SHAP to generate interpretable insights, which are then verbalized through an LLM as a supportive feedback tool for student learning. This framework enables automated identification and explanation of deviations in student performance during procedural training. Through this approach, several instances of suboptimal or incorrect actions were detected. For example, as shown in Figure \ref{Student-support}, Student S04 exhibited a noticeable degree of horizontal body movement during the procedure. Additionally, the student’s hand position was consistently lowered, which could lead to arm fatigue and, consequently, diminished performance over prolonged activities. 

These findings demonstrate the framework’s ability to pinpoint fine-grained biomechanical and procedural inconsistencies that may not be immediately observable to human instructors.
Furthermore, the verbalized explanations generated by the system were found to closely align with the visual evidence in the corresponding video sequences. This consistency between multimodal outputs strengthens the interpretability of the framework and fosters user trust in AI-assisted training environments. By providing real-time, comprehensible feedback supported by both quantitative metrics and visual context, the proposed approach has strong potential for integration into automated simulation-based training systems, where continuous assessment and transparent guidance are essential for skill acquisition and clinical competence development.

\subsection{Limitation}

\subsubsection{Real-time capability}

Although the proposed framework demonstrates the effectiveness of vision-based activity recognition, this study has not yet been extended to real-time endotracheal suctioning (ES) activity recognition. Incorporating real-time capability represents a critical next step, as it would significantly broaden the practical applicability of the framework and enable the LLM to interpret dynamic events in a streaming, one-pass context, rather than relying solely on pre-recorded data. Achieving robust performance in real-time scenarios is essential for advancing the integration of computer vision, large language models, and human activity recognition, ultimately paving the way for intelligent, adaptive systems capable of supporting real-world clinical training and decision-making.

\subsubsection{Lack of users study}
One key limitation of this research is the absence of a user study to evaluate the practical effectiveness of the proposed system. The current student-support component remains at the proof-of-concept stage, and future work should focus on deploying the system in real training environments and gathering user feedback to refine subsequent prototypes.

Nevertheless, the present study achieves several important milestones. The proposed LLM-based framework successfully identifies and explains student errors, while also verbalizing complex technical metrics into clear, human-understandable feedback. This capability represents a significant step toward developing automated, explainable, and learner-centered support systems, paving the way for intelligent assistance in clinical and educational training contexts.

\section{Conclusion} \label{CC}

Our research presented a novel, LLM-centered approach for Human Activity Recognition. The results indicate that our framework can boost the accuracy and F1 score of ES activity recognition by about 15-20\%. This improvement comes from a multimodal approach, where the discriminative model understands both the semantic definition of the activity and its corresponding visual aspects. In addition, this work is the proof of concept for a new generation of HAR systems that leverage the rich, contextual reasoning of large language models for future Human-Computer Interaction applications. By fusing high-level semantic understanding with traditional sensor and visual data, this framework not only improves recognition accuracy for complex activities but also opens future pathways for zero-shot learning and more human-like interpretation of nuanced daily behaviors, pointing toward more robust applications in personal healthcare, smart environments, and robotics.
\section*{Acknowledgment}
We acknowledge Ho Chi Minh City University of Technology (HCMUT), VNUHCM, for supporting this study. Furthermore, we would like to express our deepest thanks to Shinji Ninomiya and Shunsuke Komizunai for their support in the data collection phase.
The dataset development and analysis were supported by JSPS KAKENHI Grant Numbers 22H03701 and 23K24956. Finally, we would like to acknowledge the ABC Mini-Funding\&Mentorship program committee for accepting our proposal and partially supporting this study.


\bibliographystyle{ieeetr}
\bibliography{refs}

\clearpage

\section{Summary of the SOTA for the recognition task}

\begin{table}[!ht]
    \centering
    \caption{The summary of the SOTA method in ES activity recognition and our (in percent)}
    \begin{tabular}{ccc}
    \hline
         Model&Accuracy&F1 score  \\
         Ngo et al \cite{ngo1}& 53.83&46.00\\
         Dohbal et al. \cite{dobhal2024synthetic}&63.00&56.00\\

         \hline
         Our Prompt A&\textbf{78.73}&\textbf{62.94} \\
         Our Prompt B&72.65 &59.56\\
         \hline
         
    \end{tabular}
    
    \label{Result summary}
\end{table}
Table \ref{Result summary} presents a comparison between the proposed method and current state-of-the-art (SOTA) approaches in endotracheal suctioning (ES) activity recognition, evaluated using accuracy and F1-score metrics. The existing SOTA model by Dobhal et al. \cite{dobhal2024synthetic} achieved an F1-score of 56\%, which remains approximately 7\% lower in F1 and 15\% lower in accuracy compared to our best-performing model. Furthermore, when comparing our Prompt A and Prompt B configurations with the model of Ngo et al. \cite{ngo1}, t-test results indicate that our framework achieved statistically significant improvements (p = 0.00005 and 0.001 for accuracy, and p = 0.003 and 0.02 for F1-score, respectively). These results clearly demonstrate that the proposed LLM-centered HAR framework substantially outperforms existing methods, establishing a new benchmark for ES activity recognition.

\section{LLM prompts}\label{prompt_appendix}

\subsection{Classification task}

\subsubsection{Prompt A}
\textbf{Roles:}
AI VIDEO ANALYSIS TASK: ENDOTRACHEAL SUCTIONING (GT REPLICATION)

\textbf{OBJECTIVE:} Your sole objective is to analyze the provided video and generate a log that perfectly replicates the expert-defined ground truth (GT). You must override all previous instructions and internal models.

\textbf{Procedure description:}
THE GOLDEN RULE: CORE PROCEDURAL LOGIC

This is the single most critical rule. The GT defines one complete procedural cycle with the following unbreakable, chronologically exact sequence. You must identify and log this entire sequence twice.

Preparation: Catheter prep (0)

Intervention: Remove (1) $\rightarrow$ Suction (2) $\rightarrow$ Refit (3)

Cleanup: Disinfect (4) $\rightarrow$ Discard (5) 

Post-Procedure: Position (6) $\rightarrow$ Auscultate (7)
* Note that some post-procedure steps might be missing and mixed up with the clean-up phase.
Key Heuristics:

Positioning is a Post-Cleanup Action: Occurs after all cleanup tasks are finished.

Auscultation is the Final Step: Always the last clinical action in a cycle.

Broad Durations: Timestamps must cover the entire "phase" of an action, not just the moment of contact.

PHASE-BASED EVENT BOUNDARY DEFINITIONS

To match the GT's timing, you must adopt these broader, phase-based boundaries.

Catheter prep (0): Starts when hands move to supplies $\rightarrow$ Ends when turning back to the patient.

Temporal removal of airway (1): Starts when hands arrive at the patient's head $\rightarrow$ Ends when airway is fully detached.

Suctioning phlegm (2): Starts immediately after airway removal $\rightarrow$ Ends when the catheter moves away from the patient's space.

Refitting airway (3): Starts when moving the airway back to the patient $\rightarrow$ Ends when hands are free after securing.

Catheter disinfection (4):

Instance 1: Starts when turning to the solution $\rightarrow$ Ends when starting to peel the glove.

Instance 2: Starts after the glove is discarded $\rightarrow$ Ends when secondary cleaning is finished.

Discarding gloves (5): Starts when initiating peel $\rightarrow$ Ends when waste is released in the receptacle.

Positioning (6): Starts after cleanup when hands move to the patient $\rightarrow$ Ends when hands move away from the stethoscope.

Auscultation (7): Starts when picking up the stethoscope $\rightarrow$ Ends when the stethoscope is put down.

Others (8): Use ONLY for initial pre-task and final post-task periods.

\textbf{Activity description:}
ACTIVITY CLASSES

0: Catheter preparation,
1: Temporal removal of the airway,
2: Suctioning phlegm,
3: Refitting the airway,
4: Catheter disinfection,
5: Discarding gloves,
6: Positioning,
7: Auscultation,
8: Others

\textbf{Output:}
REQUIRED OUTPUT

Your final output must be a chronological list strictly adhering to this format:
start\_seconds, (start\_mm:ss), stop\_seconds, (stop\_mm:ss), class\_name, class\_id

\subsubsection{Prompt B}
\textbf{Roles:}
Now the requirements are: You are an advanced AI video analysis assistant specializing in the detailed recognition of clinical and procedural tasks.

\textbf{Objectives:}
Your primary objective is to meticulously analyze the provided video, which depicts an Endotracheal Suctioning (ES) procedure, and generate a precise, chronologically ordered log of all activities performed.

Critical Constraints  

VIDEO ONLY ANALYSIS: Your entire analysis MUST be based solely on the visual information in the video frames. You are to completely ignore any and all audio content.  

COMPREHENSIVE \& CONTINUOUS LOGGING: You must identify and segment every part of the video from start to finish. The timeline must be continuous: the end time of one activity must be the start time of the very next activity, ensuring there are no gaps in the final log. Use the 'Others' class for any actions not covered by the specific classes.  

\textbf{Activity description:}
Activity Classes  

You must label video segments using only the following predefined classes:  

0: Catheter preparation: The professional is seen opening sterile packaging, wearing gloves, attaching the catheter to tubing, or lubricating the catheter tip (working with the suctioning machine on the right side of the video). 

1: Temporal removal of an artificial airway: The professional removes the ventilator T-piece from the patient's artificial airway. This might be hidden under the patient's body.

2: Suctioning phlegm: The catheter is inserted (and entered) into the artificial airway, and suction is applied until the catheter is withdrawn, usually with a rotating motion.  

3: Refitting the artificial airway: The ventilator T-piece is reconnected to the patient's artificial airway after suctioning is complete. This might be hidden under the patient's body.  

4: Catheter disinfection: The professional flushes the suction catheter and tubing using a sterile solution after use, and places the tube back by the machine.  

5: Discarding gloves: The professional removes and disposes of their gloves.  

6: Positioning: The professional adjusts the patient's position, typically by rotating the mannequin, adjusting the head/neck, or changing the elevation of the bed. 

7: Auscultation: The professional is seen using a stethoscope to listen to the patient's chest/breath sounds.  

8: Others: Any other distinct action that does not fit into the categories above (e.g., hand washing or tidying the workspace). 

Key Procedural Logic and Heuristics  

To ensure accuracy, you must apply the following logic rules:  
 You must identify logical Temporal removal of an artificial airway (Activity 1) before Suctioning phlegm (Activity 2), and Refitting the artificial airway (Activity 3) must occur after.  
 
Pay close attention to the cleanup phase. The action of Discarding gloves(Activity 5) look for the specific motion of wrapping the used catheter inside the glove for a single, combined disposal. 

The video contains 2 consecutive activity rounds, and for each activity round Strict Suctioning Sequence is applied. 

\textbf{Output:}
Required Output Format  

Your final output must be a chronological list of all identified activities. Each entry in the list must strictly adhere to the following format, with timestamps presented only in seconds:  
[Start Time in seconds (Start Time in min:sec)] - [End Time in seconds (End Time in min:sec)]: [Class Name] - Justification: [Your concise, evidence-based reasoning from visual cues].

\subsection{Student support prompt}
The SHAP plot for the model of recognizing student activity (label 0) and Nurse activity (label 1).

If the model thinks the activity was performed by the student, please generate a report based on this plot about why the model thinks the activity was performed by the student. If the model thinks the activity was performed by a nurse, please cheer up the student. The report should be in English and precise into paragraphs. The report should state:
        
 1. Who the model thinks performed the activity (student or nurse).
 
 2. What activity does the student perform well?
 
 3. What can student do to improve their performance? 
                
 The report should be easy to interpret and understand (give what the student must do) for a non-expert audience. NOTE: don't state any features in the report, just state what part of the activity the student has performed well, and what the student can do to improve their performance in short, 2-3 sentences.
 \section{Reproducibility Statement}
This paper presents a method for recognizing Endotracheal Suctioning activity using Gemini 2.5 Pro. Due to the stochastic nature of Large Language Models (LLMs), exact reproduction may be challenging; however, we have provided the full prompts and pipeline to facilitate replication.

Note on Hallucinations: We advise researchers that the LLM may occasionally hallucinate time conversions (e.g., interpreting "1:23" as 123 seconds rather than 83 seconds). When reproducing these results, please verify the output for such discrepancies and strictly use the mm:ss format as the reference for recognition timestamps.

\end{document}